\documentclass[10pt,twocolumn,letterpaper]{article}
\usepackage{cvpr}
\usepackage{times}
\usepackage{epsfig}
\usepackage{graphicx}
\usepackage{amsmath}
\usepackage{amssymb}
\usepackage[breaklinks=true,bookmarks=false]{hyperref}
\usepackage{array}
\usepackage{tabularx,ragged2e,booktabs,caption}
\usepackage{cite}
\usepackage{comment}
\newcolumntype{C}[1]{>{\Centering}m{#1}}

\newcolumntype{S}[1]{>{\hsize=.2\hsize}m{#1}}

\cvprfinalcopy 
\def\cvprPaperID{3503} 

\def\myparagraph#1{\medskip\noindent{\bf #1}~~}

\ifcvprfinal\fi

\begin{document}
\title{
 Dual Residual Networks Leveraging the Potential of Paired Operations\\ for Image Restoration
 }

\author{Xing Liu\textsuperscript{\textdagger} ~~~~Masanori Suganuma\textsuperscript{\textdagger\textdaggerdbl} ~~~~Zhun Sun\textsuperscript{\textdaggerdbl} ~~~~Takayuki Okatani\textsuperscript{\textdagger\textdaggerdbl} \\
\textsuperscript{\textdagger}Graduate School of Information Sciences, Tohoku University ~~~~~~~~~\textsuperscript{\textdaggerdbl}RIKEN Center for AIP \\
{\tt\small \{ryu,suganuma,zhun,okatani\}@vision.is.tohoku.ac.jp}
}
\maketitle
\thispagestyle{empty}

\begin{abstract}
In this paper, we study design of deep neural networks for tasks of image restoration. 
We propose a novel style of residual connections dubbed ``dual residual connection'', which exploits the potential of paired operations, e.g., up- and down-sampling or convolution with large- and small-size kernels. We design a modular block implementing this connection style; it is equipped with two containers to which arbitrary paired operations are inserted. Adopting the ``unraveled'' view of the residual networks proposed by Veit et al., 
we point out that a stack of the proposed modular blocks allows the first operation in a block interact with the second operation in {\em any} subsequent blocks. Specifying the two operations in each of the stacked blocks, 
we build a complete network for each individual task of image restoration. We experimentally evaluate the proposed approach on five image restoration tasks using nine datasets. The results show that the proposed networks with properly chosen paired operations outperform previous methods on almost all of the tasks and datasets.
\end{abstract}

\section{Introduction}
The task of restoring the original image from its degraded version, or image restoration, has been studied for a long time in the fields of image processing and computer vision. As in many other tasks of computer vision, the employment of deep convolutional networks have made significant progress. In this study, aiming at further improvements, we pursue better architectural design of networks, particularly the design that can be shared across different tasks of image restoration.
In this study, we pay attention to the effectiveness of paired operations on various image processing tasks. In \cite{dualupdown}, 
it is shown that a CNN iteratively performing a pair of up-sampling and down-sampling contributes to performance improvement for image-superresolution. In \cite{god_suganuma}, the authors employ evolutionary computation to search for a better design of convolutional autoencoders for several tasks of image restoration, showing that network structures repeatedly performing a pair of convolutions with a large- and small-size kernels (\eg, a sequence of conv. layers with kernel size 3, 1, 3, 1, 5, 3, and 1) perform well for image denoising. In this paper, we will show further  examples for other image restoration tasks. 
\begin{figure}[bt]
\centering
\includegraphics[width=0.75\columnwidth]{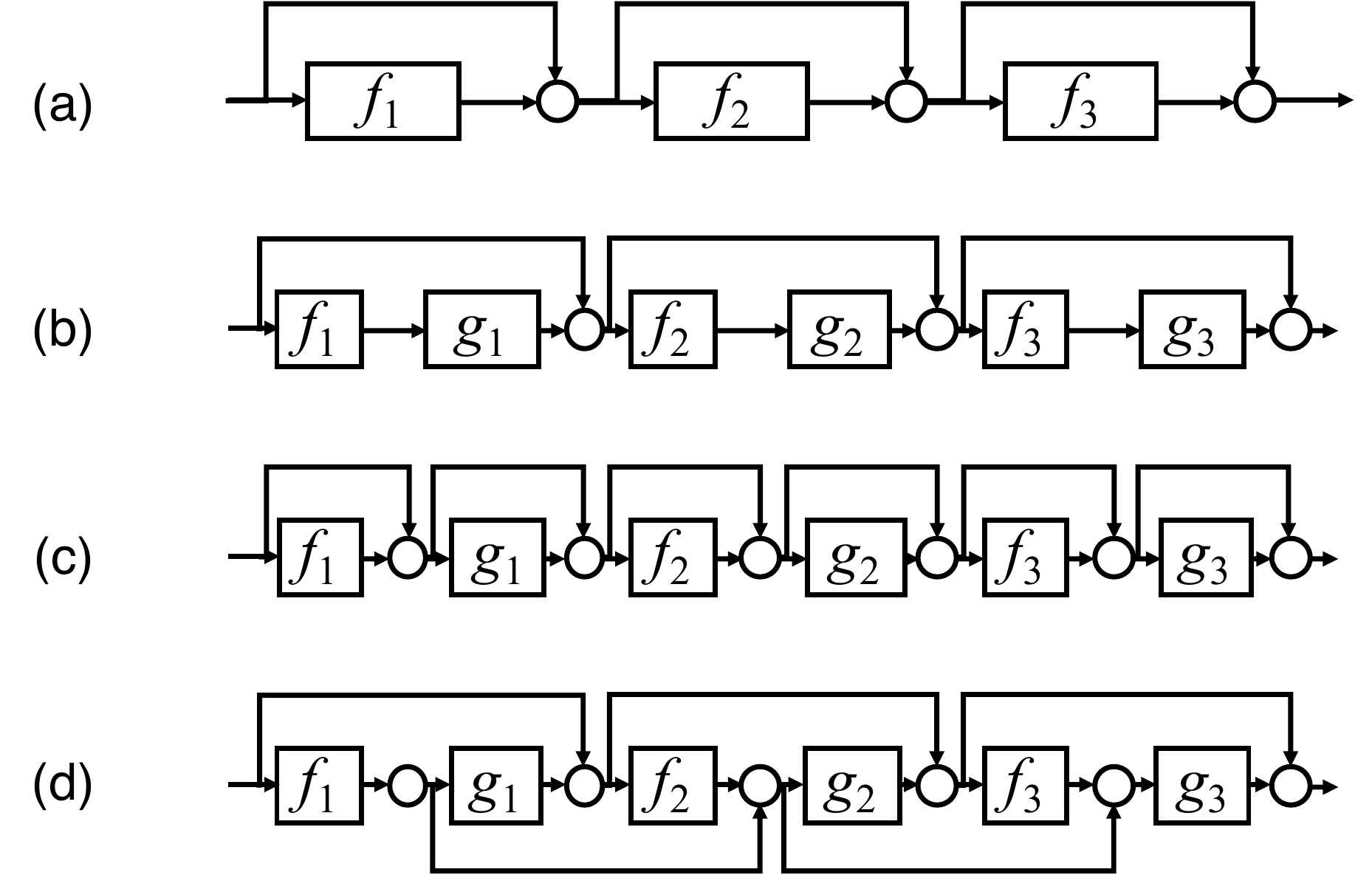}
\caption{Different construction of residual networks with a single or double basic modules. The proposed ``dual residual connection'' is (d). }
\label{fig:diagram}
\vspace{-0.3cm}
\end{figure}
Assuming the effectiveness of such repetitive paired operations, we wish to implement them in deep networks to exploit their potential. We are specifically interested in how to integrate them with the structure of residual networks. The basic structure of residual networks is shown in Fig.~\ref{fig:diagram}(a), which have become an indispensable component for the design of modern deep neural networks. 
There have been several explanations for the effectiveness of the residual networks. A widely accepted one is the ``unraveled'' view proposed by Veit \etal \cite{res_behave}: a sequential connection of $n$ residual blocks is regarded as an ensemble of many sub-networks corresponding to its implicit $2^n$ paths. A network of three residual blocks with modules $f_1$, $f_2$, and $f_3$, shown in Fig.~\ref{fig:diagram}(a), has $(2^3=)8$ implicit paths from the input to output, i.e.,  $f_1\rightarrow f_2\rightarrow f_3$, $f_1\rightarrow f_2$, $f_1\rightarrow f_3$, $f_2\rightarrow f_3$, $f_1$, $f_2$, $f_3$, and $1$. Veit \etal also showed that each block works as a computational unit that can be attached/detached to/from the main network  with minimum performance loss. Considering such a property of residual networks, how should we use residual connections for paired operations? Denoting the paired operations by $f$ and $g$, the most basic construction will be to treat $(f_i, g_i)$ as a unit module, as shown in Fig.~\ref{fig:diagram}(b). In this connection style, $f_i$ and $g_i$ are always paired for any $i$ in the possible paths. In this paper, we consider another connection style shown in Fig.~\ref{fig:diagram}(d), dubbed ``dual residual connection''. This style enables to pair $f_i$ and $g_j$ for any $i$ and $j$ such that $i\leq j$. 
In the example of Fig.\ref{fig:diagram}(d), all the combinations of the two operations, $(f_1,g_1)$, $(f_2,g_2)$, $(f_3,g_3)$, $(f_1,g_2)$, $(f_1,g_3)$, and $(f_2,g_3)$, emerge in the possible paths. We conjecture that this increased number of potential interactions between $\{f_i\}$ and $\{g_j\}$ will contribute to improve performance for image restoration tasks. Note that it is guaranteed that $f_\cdot$ and $g_\cdot$ are always paired in the possible paths. This is not the case with other connection styles such as the one depicted in Fig.~\ref{fig:diagram}(c). 
We call the building block for implementing the proposed dual residual connections {\em Dual Residual Block} (DuRB); see Fig.~\ref{fig:face}. We examine its effectiveness on {five image restoration tasks} shown in Fig.~\ref{fig:face} using {nine datasets}. DuRB is a generic structure that has two containers for the paired operations, and the users choose two operations for them. For each task, we specify the paired operations of DuRBs as well as the entire network.
Our experimental results show that our networks outperform the state-of-the-art methods in these tasks, which supports the effectiveness of our approach.

\begin{figure}[!t]
\centering
\includegraphics[clip,width=8cm]{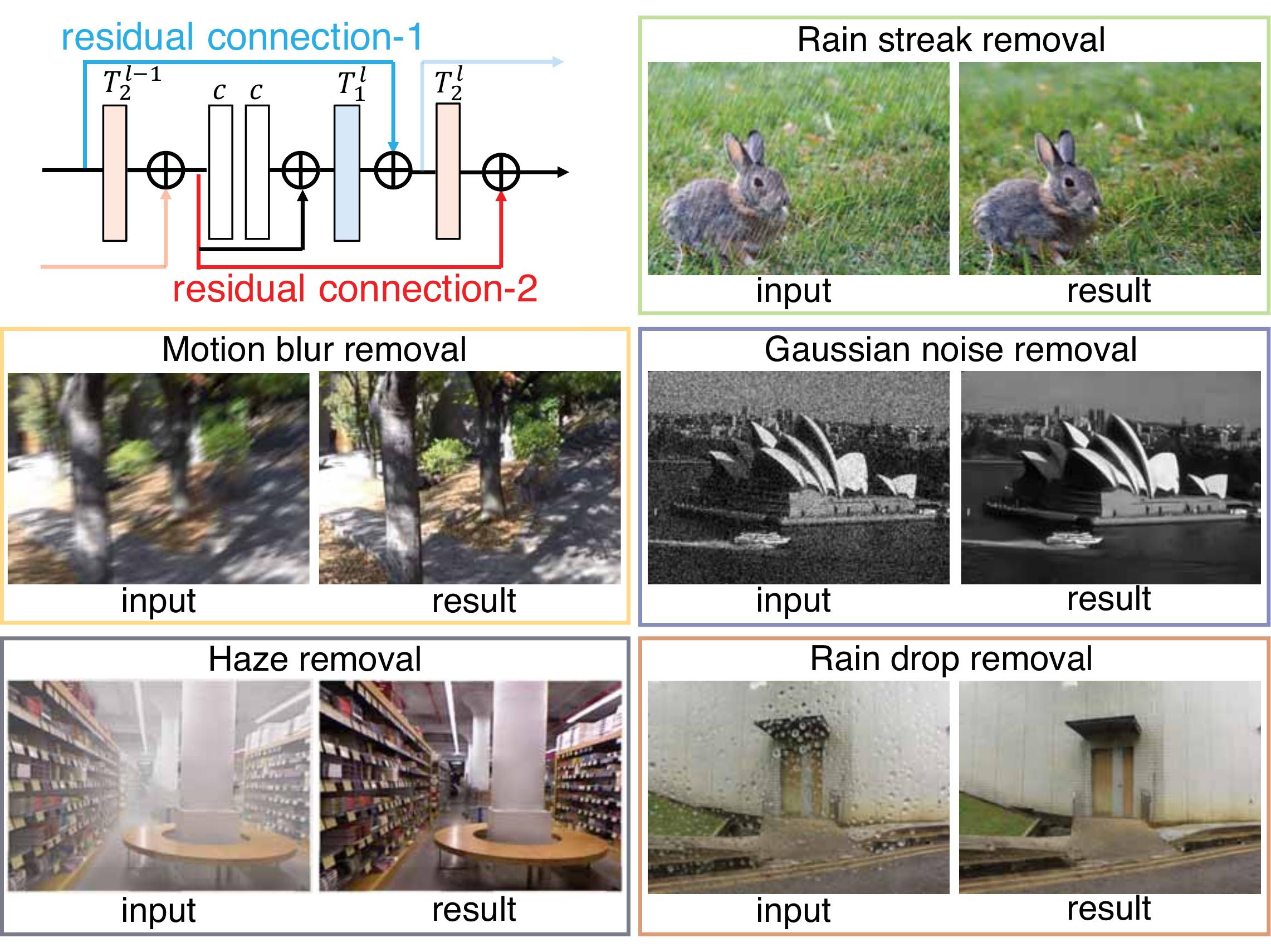}
\caption{Upper-left: the structure of a unit block having the proposed dual residual connections; $T^{l}_{1}$ and $T^{l}_{2}$ are the containers for two paired operations; $c$ denotes a convolutional layer. Other panels: five image restoration tasks considered in this paper. }
\label{fig:face}
\vspace{-0.3cm}
\end{figure}

\section{Related Work}
\myparagraph{Gaussian noise removal} 
Application of 
neural networks to noise removal has a long history \cite{cnn_denoise_old1,cnn_denoise_old2,cnn_denoise_old3,dncnn,ffdnet}. 
Mao \etal \cite{REDnet} proposed REDNet, which consists of multiple convolutional and de-convolutional layers with symmetric skip connections over them.
Tai \etal \cite{Memnet} proposed MemNet with local memory blocks and global dense connections, showing that it performs better than REDNet. 
However, Suganuma \etal \cite{god_suganuma} showed that standard convolutional autoencoders with repetitive pairs of convolutional layers with large- and small-size kernels outperform them by a good margin, which are found by architectural search based on evolutionary computation. 

\myparagraph{Motion blur removal}
This task has a long history of research. Early works  \cite{Fergus,xu-blur,deblur3,deblur4} attempt to simultaneously estimate both blur kernels and sharp images. 
Recently, CNN-based methods \cite{sun-blur,gong-blur,nah,DeBlurGAN,repeat_ed} achieve good performance for this task. Nah \etal \cite{nah} 
proposed a coarse-to-fine approach along with a modified residual block \cite{resnet}.
Kupyn \etal \cite{DeBlurGAN} proposed an approach based on  Generative Adversarial Network (GAN) \cite{GAN}. New datasets were created in \cite{nah} and \cite{DeBlurGAN}.

\myparagraph{Haze removal} 
Many studies assume the following model of haze: $I(x) = J(x)t(x) + A(x)(1 - t(x))$, where $I$ denotes a hazy scene image, $J$ is the true scene radiance (the clear image), $t$ is a transmission map, $A$ is global atmospheric light. 
The task is then to estimate $A$, $t$, and thus $J(x)$ from the input $I(x)$  \cite{kaiming_dehaze,dehaze_relate1,dehaze_relate2,dehaze_zhanghe,Yang_2018_ECCV}.  Recently, Zhang \etal \cite{dehaze_zhanghe} proposed a method that uses CNNs to jointly estimate $t$ and $A$, which outperforms previous approaches by a large margin. Ren \etal \cite{Ren_dehaze} and Li \etal \cite{DehazeGAN} proposed method to directly estimate $J(x)$ without explicitly estimating $t$ and $A$. Yang \etal \cite{Yang_2018_ECCV} proposed a method that integrates CNNs to classical prior-based method.

\myparagraph{Raindrop detection and removal} Various approaches \cite{raindrop-relate1,raindrop-relate2,raindrop-relate3,raindrop-relate4,raindrop-relate5} have been proposed to tackle this problem in the literature. Kurihata \etal \cite{raindrop-relate1} proposed to detect raindrops with raindrop-templates learned using PCA. Ramensh \cite{raindrop-relate3} proposed a method based on K-Means clustering and median filtering to estimate clear images. Recently, Qian \etal \cite{raindrop18} proposed a hybrid network consisting of a convolutional-LSTM for localizing raindrops and a CNN for generating clear images, which is trained in a GAN framework. 

\myparagraph{Rain-streak removal}
Fu \etal \cite{xiamen} use ``guided image filtering'' \cite{guided_image_filter} to extract high-frequency components of an image, and use it to train a CNN for rain-streak removal.
Zhang \etal \cite{derain_zhanghe} proposed to jointly estimate rain density and de-raining result to alleviate the non-uniform rain density problem. Li \etal \cite{rescan} regards a heavy rainy image as a clear image added by an accumulation of multiple rain-streak layers and proposed a RNN-based method to restore the clear image. Li \etal \cite{derain_sota_acm} proposed an non-locally enhanced version of DenseBlock \cite{densenet} for this task, their network outperforms previous approaches by a good margin.

\begin{table*}[t]
 \caption{Performance of the three connection types of Fig.~1(b)-(c). `-'s indicate infeasible applications.}
 \vspace{-0.3cm}
 \label{tab:denoise_results0} 
 \centering 
 \begin{tabular}{|c|c|c|c|c|c|c|}
 \hline
  & Gaussian noise & Real noise & Motion blur & Haze & Raindrop & Rain-streak\\
 \hline
  (b) & 24.92 / 0.6632 & 36.76 / 0.9620 & 29.46 / 0.9035& 31.20 / 0.9803& 24.70 / 0.8104 & 32.85 / 0.9214\\
  (c) & 24.85 / 0.6568 & 36.81 / 0.9627 & -/-& -/-&  25.12 / 0.8151 & 33.13 / 0.9222\\
  (d) & {\bf25.05} / {\bf0.6755} & {\bf36.84} / {\bf0.9635} & {\bf29.90} / {\bf0.9100}& {\bf32.60} / {\bf0.9827}& {\bf 25.32} / {\bf 0.8173} & {\bf 33.21} / {\bf 0.9251}\\
 \hline
 \end{tabular}
\end{table*}

\section{Dual Residual Blocks}
\label{sec:FourDuRBs}
The basic structure of the proposed Dual Residual Block (DuRB) is shown in the upper-left corner of Fig.~\ref{fig:face}, in which we use $c$ to denote a convolutional layer (with $3 \times 3$ kernels) and $T^{l}_1$ and $T^{l}_2$ to denote the containers for the paired first and second operations, respectively, in the $l^{th}$ DuRB in a network. Normalization layers (such as batch normalization \cite{batchnorm} or instance normalization \cite{instancenorm}) and ReLU \cite{relu} layers can be incorporated when it is necessary.
\begin{figure}[!t]
\centering
\includegraphics[width=.8\columnwidth]{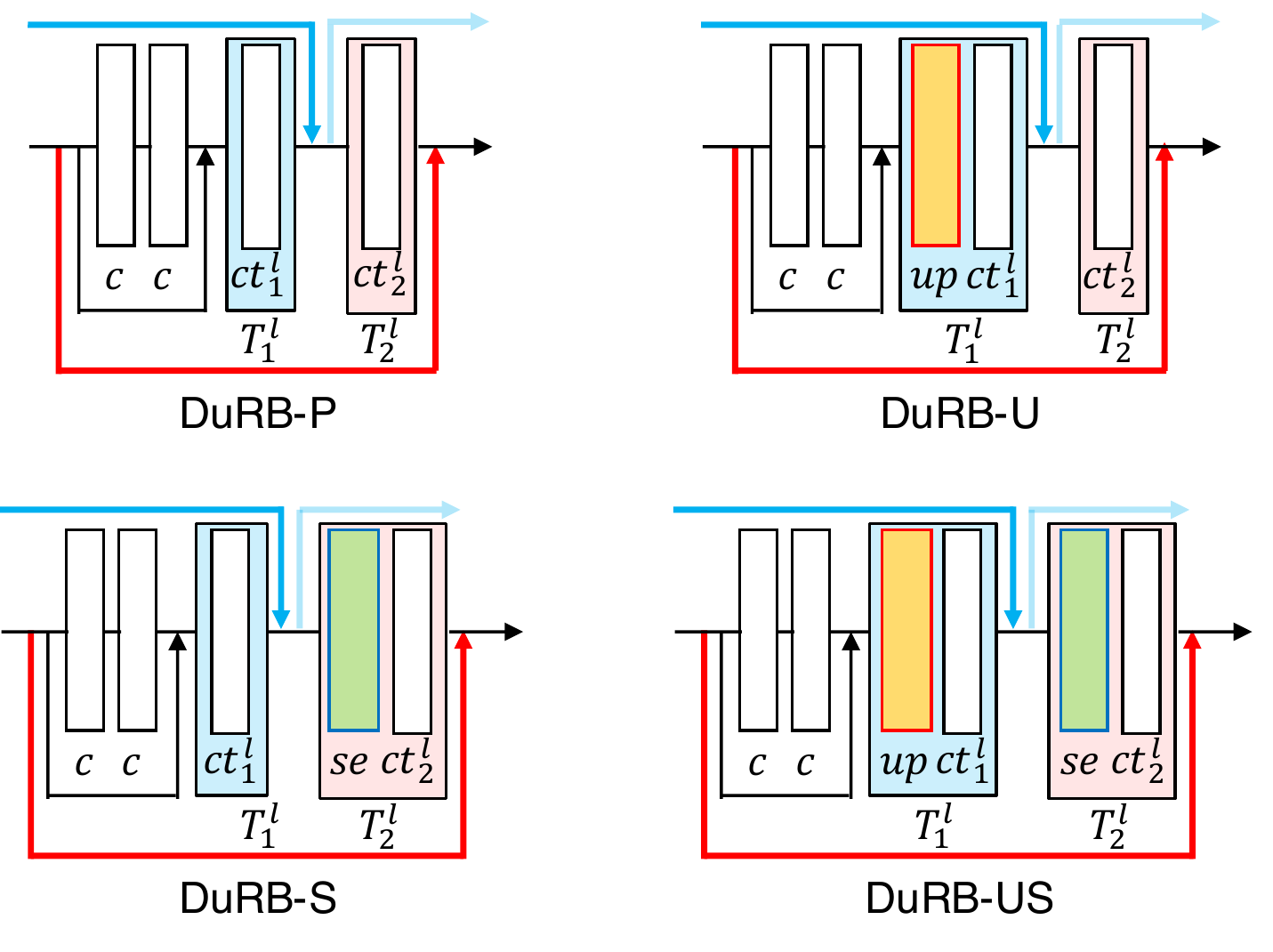}
\caption{Four different implementations of the DuRB; $c$ is a convolutional layer with $3 \times 3$ kernels; $ct^{l}_1$ and $ct^{l}_2$ are convolutional layers, each with kernels of a specified size and dilation rate; $up$ is up-sampling (we implemented it using PixelShuffle \cite{pixelshuffle}); 
$se$ is SE-ResNet Module \cite{senet} that is in fact a channel-wise attention mechanism.}
\label{fig:Four_DuRBs}
\vspace{-0.3cm}
\end{figure}
We design DuRBs for each individual task, or equivalently choose the two operations to be inserted into the containers $T_1^l$ and $T_2^l$. We will use four different designs of DuRBs, DuRB-P, DuRB-U, DuRB-S, and DuRB-US, which are shown in Fig.~\ref{fig:Four_DuRBs}. The specified operations for [$T_1^l$, $T_2^l$] are [conv., conv.] for DuRB-P, [up-sampling+conv., 
down-sampling (by conv. with stride=2)] for DuRB-U, [conv., channel-wise attention\footnote{It is implemented using the SE-ResNet Module \cite{senet}.}+conv.] for DuRB-S, and [up-sampling+conv., channel-wise attention+down-sampling] for DuRB-US, respectively. We will use DuRB-P for noise removal and raindrop removal, DuRB-U for motion blur removal, DuRB-S for rain-streak and raindrop removal, and DuRB-US for haze removal.

Before proceeding to further discussions, we present here experimental results that show the superiority of the proposed dual residual connection to other connection styles shown in Fig.~\ref{fig:diagram}(b) and (c). In the experiments, three networks build on the three base structures (b), (c), and (d) of Fig.~\ref{fig:diagram} were evaluated on the five tasks.
For Gaussian\&real-world noise removal, motion blur removal, haze removal, raindrop and rain-streak removal, we use DuRB-P, DuRB-U, DuRB-US, DuRB-S\&DuRB-P and DuRB-S to construct the base structures. Number of blocks and all the operations in the three structures as well as other experimental configurations are fixed in each comparison. The datasets for the six comparisons are BSD-grayscale, Real-World Noisy Image Dataset, GoPro Dataset, Dehaze Dataset, RainDrop Dataset and DID-MDN Data.
Table \ref{tab:denoise_results0} shows their performance. Note that `-' in the table indicate that the connection cannot be applied to DuRB-U and DuRB-US due to the difference in size between the output of $f$ and the input to $g$. It can be seen that the proposed structure (d) performs the best for all the tests. 

\section{Five Image Restoration Tasks}
In this section, we describe how the proposed DuRBs can be applied to multiple image restoration tasks, noise removal, motion blur removal, haze removal, raindrop removal and rain-streak removal.

\begin{figure}[h]
\centering
\includegraphics[clip, width=\columnwidth]{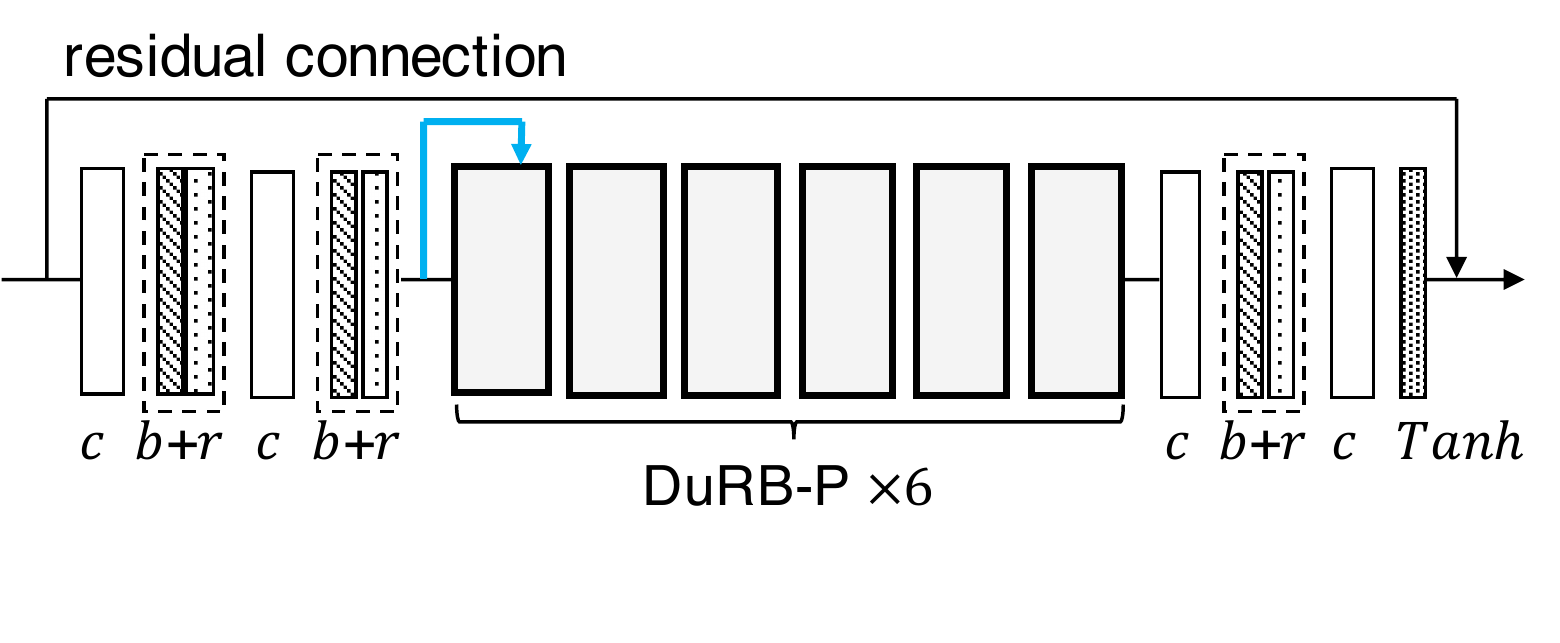}
\vspace{-1.2cm}
\caption{DuRN-P: dual residual network with DuRB-P's [conv. w/ a large kernel and conv. w/ a small kernel] for Gaussian noise removal.
$b+r$ is a batch normalization layer followed by a ReLU layer; and $Tanh$ denotes hyperbolic tangent function.
}
\label{fig:ClassicDuRN}
\vspace{-0.3cm}
\end{figure}

\subsection{Noise Removal}
\label{sec:noise}

\myparagraph{Network Design}
We design the entire network as shown in Fig.~\ref{fig:ClassicDuRN}. It consists of an input block, the stack of six DuRBs, and an output block, additionally with an outermost residual connection from the input to output. The layers $c$, $b+r$ and $Tanh$ in the input and output blocks are convolutional layer (with 3$\times$3 kernels, stride $=1$), batch normalization layer followed by a ReLU layer, and hyperbolic tangent function layer, respectively. 

\begin{figure}[!t]
\centering
\includegraphics[clip, width=.9\columnwidth]{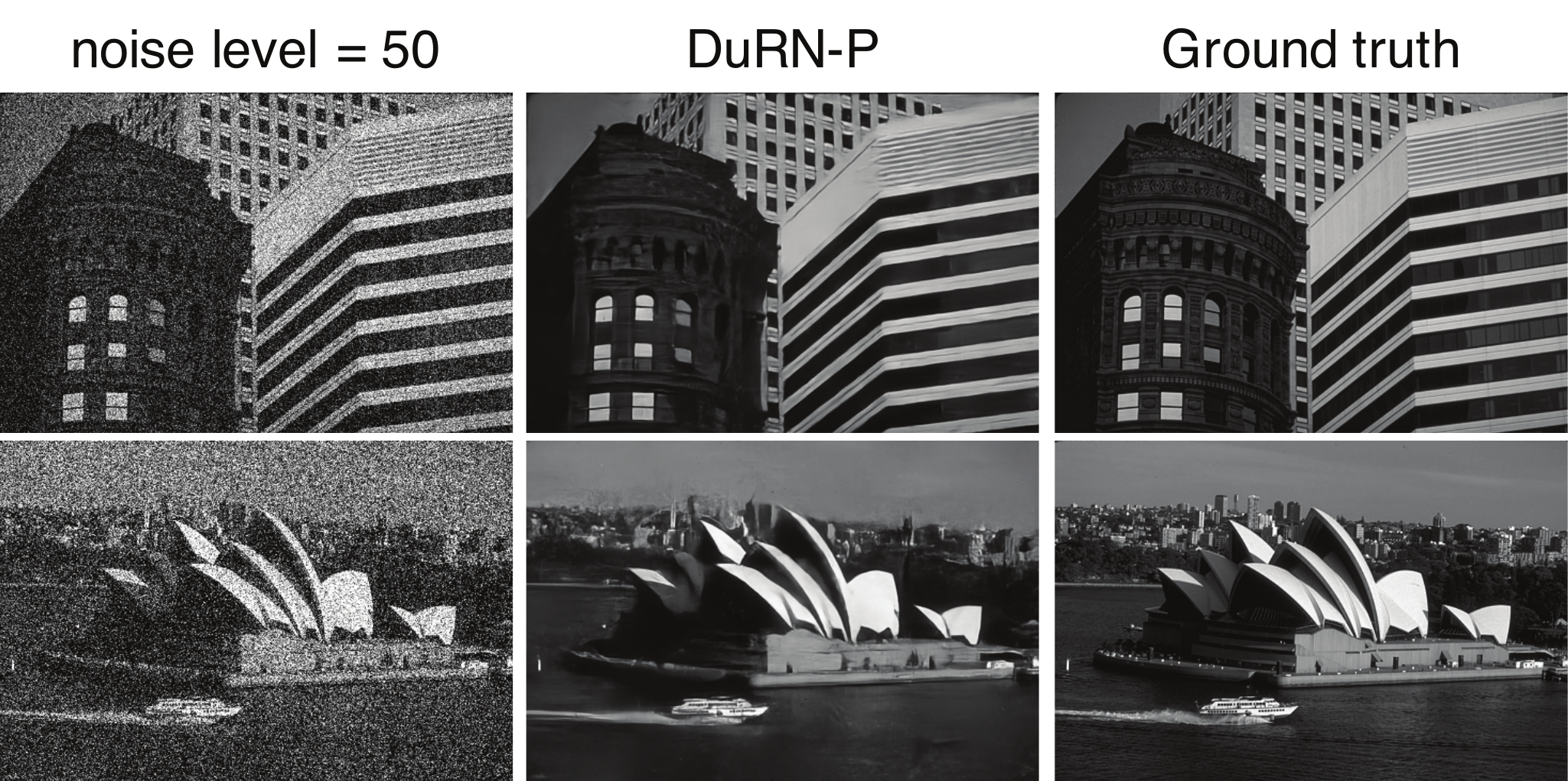}
\caption{Some examples of the results by the proposed DuRN-P for additive Gaussian noise removal. Sharp images can be restored from heavy noises ($\sigma = 50$). }
\label{fig:BSD_examples}
\vspace{-0.3cm}
\end{figure}

\begin{figure}[!t]
\centering
\includegraphics[clip, width=.9\columnwidth]{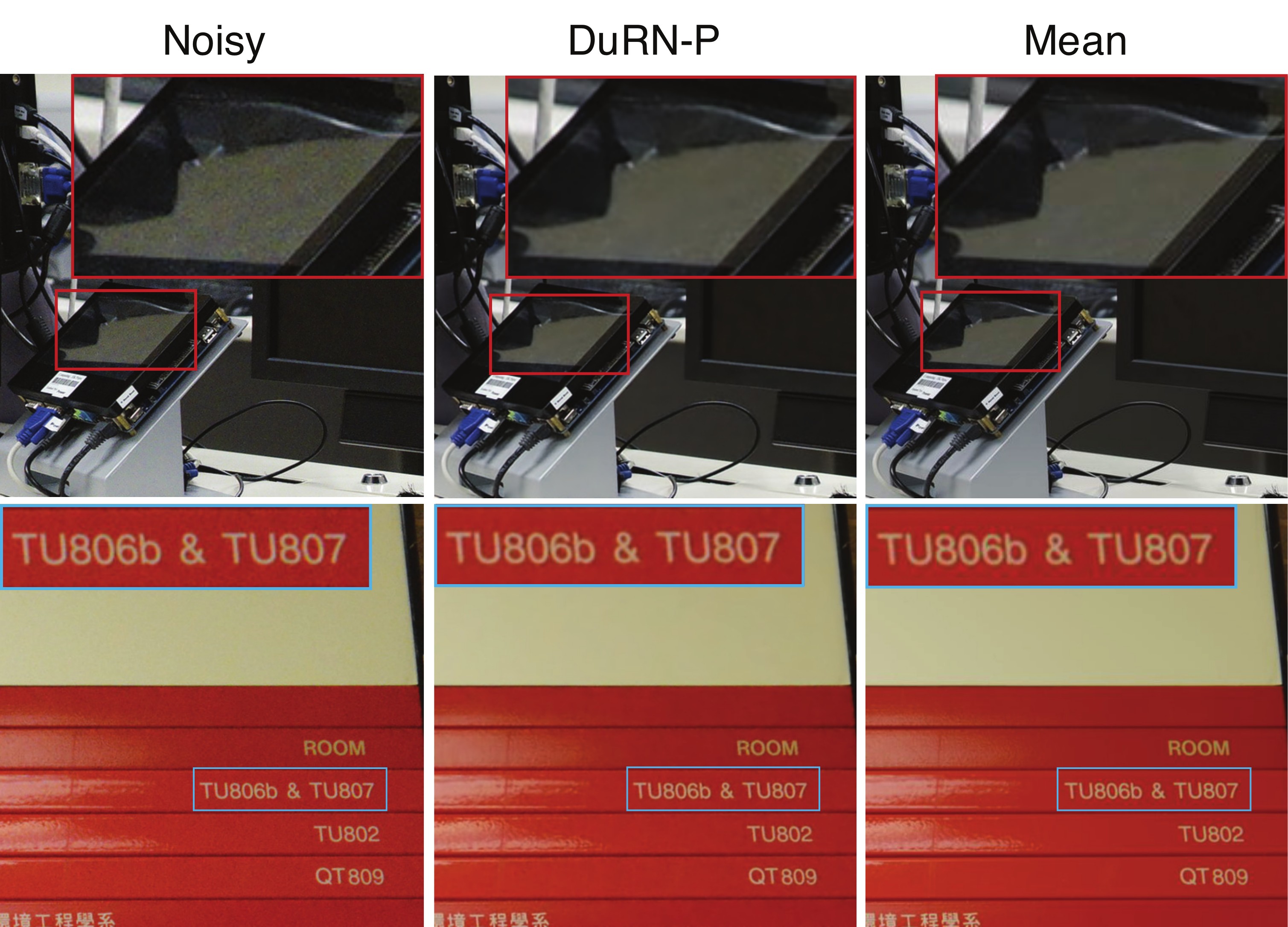}
\caption{Examples of noise removal by the proposed DuRN-P for images from Real-World Noisy Image Dataset. The results are sometimes even better than the mean image (used as the ground truth); see the artifact around the letters in the bottom.} 
\label{fig:HKNoise_results}
\vspace{-0.3cm}
\end{figure}
We employ DuRB-P (i.e., the design in which each of the two operations is single convolution; see Fig.~\ref{fig:Four_DuRBs}) for DuRBs in the network. Inspired by the networks discovered by neural architectural search for noise removal in \cite{god_suganuma},  we choose for $T_1$ and $T_2$ convolution with large- and small-size receptive fields.
We also choose the kernel size and dilation rate for each DuRB so that the receptive field of convolution in each DuRB grows its size with $l$. More details are given in the supplementary material.
We set the number of channels to 32 for all the layers. We call the entire network DuRN-P. For this task, we employed $l_2$ loss for training the DuRN-P.

\myparagraph{Results: Additive Gaussian Noise Removal}
We tested the proposed network on the task of removing additive Gaussian noise of three levels (30, 50, 70) from a gray-scale noisy image. Following the same experimental protocols used by previous studies, we trained and tested the proposed DuRN-P using the training and test subsets (300 and 200 grayscale images) of the BSD-grayscale dataset \cite{BSD}. More details of the experiments are provided in the supplementary material. We show the quantitative results in Table \ref{tab:denoise_results} and qualitative results in Fig.~\ref{fig:BSD_examples}. It is observed from Table \ref{tab:denoise_results} that the proposed network outperforms the previous methods for all three noise levels.

\begin{table}[t]
\caption{Results for additive Gaussian noise removal on BSD200-grayscale and noise levels (30, 50, 70). The numbers are PSNR/SSIM.}
\vspace{-0.3cm}
\label{tab:denoise_results}
\resizebox{\linewidth}{!}{
\begin{tabular}{|c||c|c|c|c|}
 \hline
  & 30 & 50& 70\\
 \hline
  \small{REDNet\cite{REDnet}}& \small{27.95 / 0.8019}& \small{25.75 / 0.7167}& \small{24.37 / 0.6551} \\ 
  \small{MemNet\cite{Memnet}}& \small{28.04 / 0.8053}& \small{25.86 / 0.7202}& \small{24.53 / 0.6608} \\  
  \small{E-CAE\cite{god_suganuma}}& \small{28.23 / 0.8047} & \small{26.17 / 0.7255} & \small{24.83 / 0.6636} \\  
  \hline\hline
  \textbf{\small{DuRN-P (ours)}}& \small{{\bf 28.50} / {\bf 0.8156}}& \small{{\bf 26.36} / {\bf 0.7350}} & \small{{\bf 25.05} / {\bf 0.6755}} \\
 \hline
\end{tabular}}
\end{table}

\myparagraph{Results: Real-World Noise Removal}
We also tested the DuRN-P on the Real-World Noisy Image Dataset \cite{HKRealNoise}, which consists of 40 pairs of an instance image (a photograph taken by a CMOS camera) and the mean image (mean of multiple shots of the same scene taken by the CMOS camera). We removed all the batch normalization layers from the DuRN-P for this experiment, as the real-world noise captured in this dataset do not vary greatly. The details of the experiments are given in the supplementary material. The quantitative results of three previous methods and our method are shown in Table \ref{tab:HKNoise_results}. 
We used the authors' code to evaluate the three previous methods. (As the MemNet failed to produce a competitive result, we left the cell empty for it in the table.) It is seen that our method achieves the best result despite the smaller number of parameters. 
Examples of output images are shown in Fig.~\ref{fig:HKNoise_results}. We can observe that the proposed DuRN-P has cleaned noises well. It is noteworthy that the DuRN-P sometimes provides better images than the ``ground truth'' mean image; see the bottom example in Fig.~\ref{fig:HKNoise_results}. 

 \begin{table}[t]
 \caption{Results on the Real-World Noisy Image Dataset \cite{HKRealNoise}. The results were measured by PSNR/SSIM. The last row shows the number of parameters for each CNN.}
 \vspace{-0.3cm}
 \label{tab:HKNoise_results}
 \resizebox{\linewidth}{!}{
 \begin{tabular}{|c|c|c|c|c|}
  \hline
   &REDNet\cite{REDnet} &MemNet\cite{Memnet} &E-CAE\cite{god_suganuma} &\textbf{DuRN (ours)} \\
  \hline
   PSNR/SSIM& {35.56} / 0.9475& - / -& 35.45 / {0.9492}& {\bf 36.83} / {\bf 0.9635} \\
  \hline
   \# of param.& $4.1 \times 10^{6}$& $2.9 \times 10^{6}$& $1.1 \times 10^{6}$& $8.2 \times 10^{5}$\\
  \hline
 \end{tabular}}
 \end{table}

\subsection{Motion Blur Removal}
\label{sec:motion-blur}

\begin{figure*}[!t]
\centering
\includegraphics[clip, width=\textwidth]{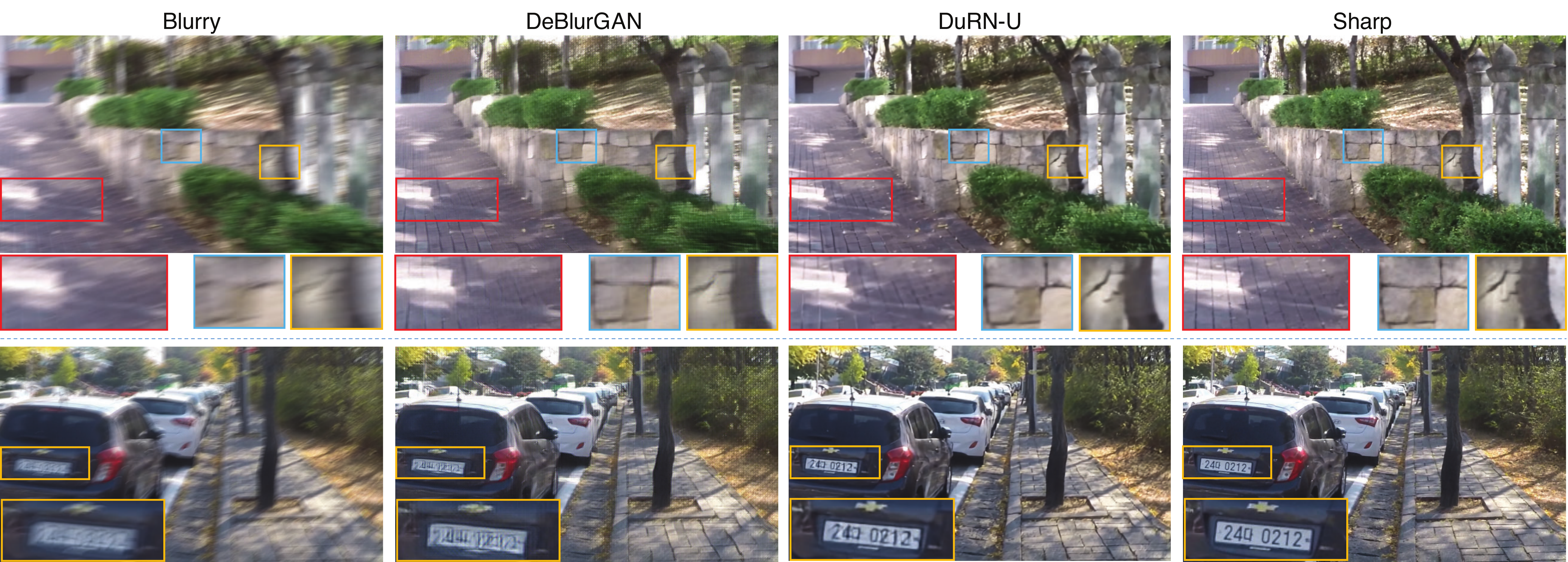}
\caption{Examples of motion blur removal on GoPro-test dataset. }
\label{fig:go_pro_vis}
\vspace{-0.3cm}
\end{figure*}

\begin{figure}[!t]
\centering
\includegraphics[clip, width=\columnwidth]{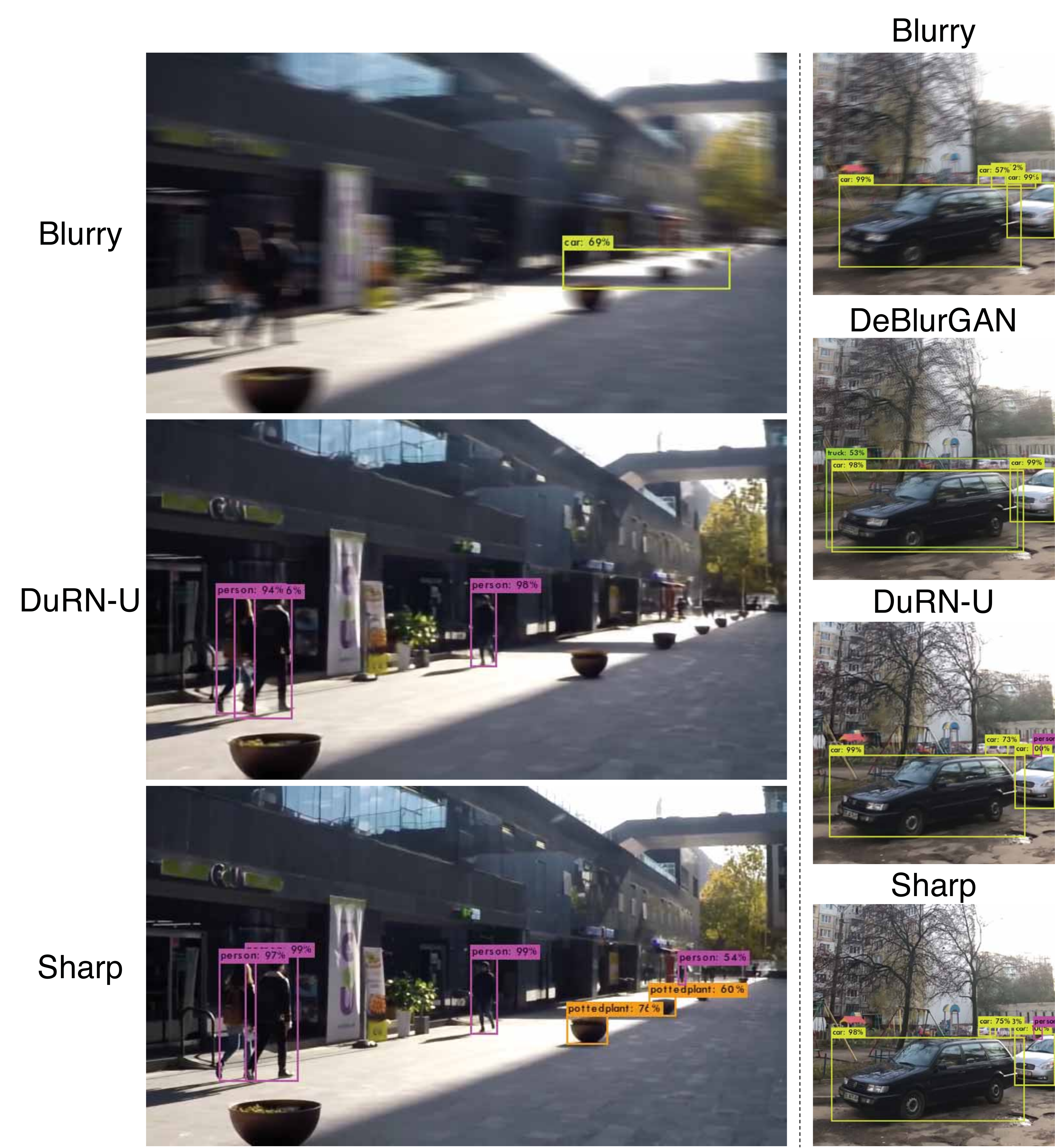}
\caption{Examples of object detection from original blurred images and their deblurred versions. }
\label{fig:the_car_det}
\vspace{-0.3cm}
\end{figure}

\begin{figure}[t]
\centering
\includegraphics[clip, width=\columnwidth]{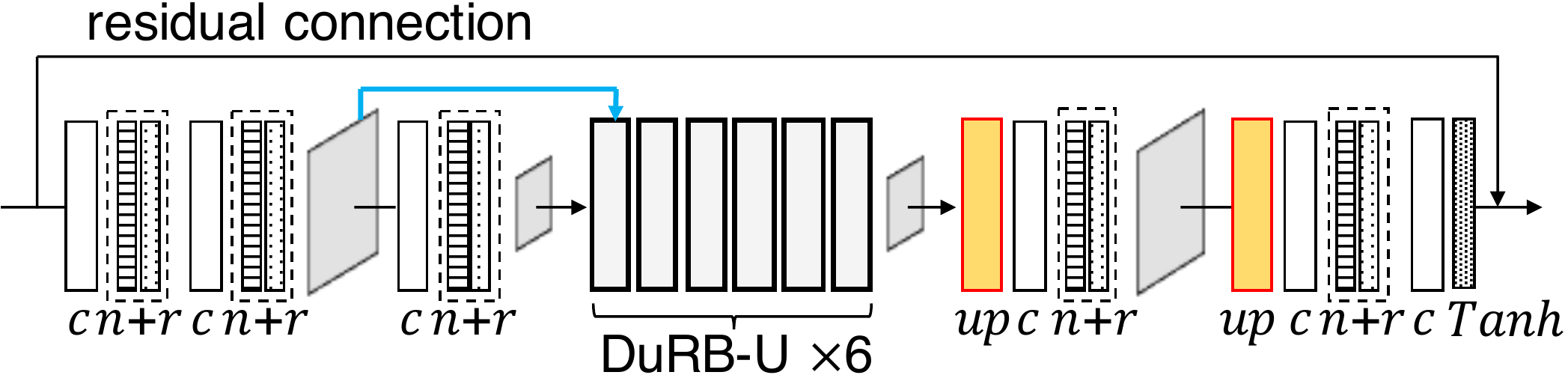}
\caption{DuRN-U: Dual Residual Network with DuRB-U's  (up- and down-sampling) for motion blur removal. $n+r$ denotes an instance normalization layer followed by a ReLU layer. }
\label{fig:The11_PixShuf}
\end{figure}

The task is to restore a sharp image from its motion blurred version without knowing the latent blur kernels (i.e., the ``blind-deblurring'' problem). 

\myparagraph{Network Design}
Previous works such as \cite{repeat_ed} reported that the employment of up- and down-sampling operations is effective for this task. Following this finding, we employ up-sampling and down-sampling for the paired operation. We call this as DuRB-U; see Fig.~\ref{fig:Four_DuRBs}. 
We use PixelShuffle \cite{pixelshuffle} for implementing up-sampling. For the entire network design, following many previous works \cite{repeat_ed,DeBlurGAN,dehaze_zhanghe,cyclegan}, we choose a symmetric encoder-decoder network; see Fig.~\ref{fig:The11_PixShuf}. The network consists of the initial block, which down-scales the input image by 4:1 down-sampling with two convolution operations ($c$) with stride $=2$, and instance normalization + ReLU ($n+r$), and six repetitions of DuRB-U's, and the final block which up-scales the output of the last DuRB-U by applications of 1:2 up-sampling ($up$) to the original size. We call this network DuRN-U. For this task, we employed a weighted sum of SSIM and $l_1$ loss for training the DuRN-U. The details are given in the supp. material.

\begin{table}[!t]
\centering
\caption{Results of motion blur removal for the GoPro-test dataset.}
\vspace{-0.3cm}
\label{tab:go_pro_results}
\small
\begin{tabular}{|c|c|}
\hline
\multicolumn{2}{|c|}{GoPro-test} \\
\hline 
Sun \etal\cite{sun-blur} & 24.6 / 0.84 \\
Nah \etal\cite{nah} & {28.3} / {0.92} \\
Xu \etal\cite{xu-blur} & 25.1 / 0.89 \\
DeBlurGAN\cite{DeBlurGAN} & 27.2 / {\bf 0.95} \\
\hline \hline
\textbf{DuRN-U (ours)} & {\bf 29.9} / 0.91 \\
\hline
\end{tabular}
\end{table}

\begin{table}[!h]
\centering
\caption{Accuracy of object detection from deblurred images obtained by DeBlurGAN \cite{DeBlurGAN} and the proposed DuRN-U on Car Dataset. }
\vspace{-0.3cm}
\label{tab:the_car_results}
\resizebox{\linewidth}{!}{
\begin{tabular}{|c|c|c|c|}
 \hline
  & Blurred &DeBlurGAN\cite{DeBlurGAN}  &\textbf{DuRN-U (ours)}\\
 \hline
  mAP ($\%$)& 16.54& 26.17 & {\bf 31.15} \\
 \hline
\end{tabular}}
\end{table}

\begin{figure*}[t]
\centering
\includegraphics[clip, width=\textwidth]{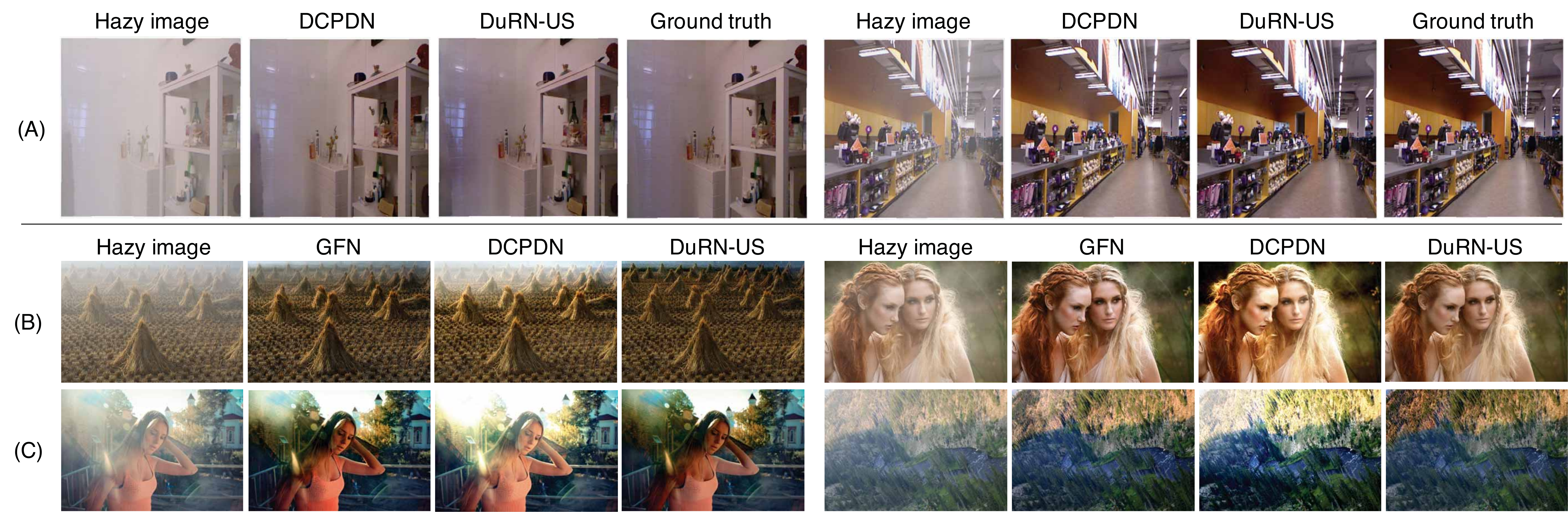}
\caption{ Examples of de-hazing results obtained by DuRN-US and others on (A) synthesized images, (B) real images and (C) light hazy images.}
\label{fig:haze_testA}
\vspace{-0.3cm}
\end{figure*}

\myparagraph{Results: GoPro Dataset}
We tested the proposed DuRN-U on the GoPro-test dataset \cite{nah} and compared its results with the state-of-the-art DeblurGAN\footnote{The DeblurGAN refers the ``DeblurGAN-wild'' introduced in the original paper \cite{DeBlurGAN}.} \cite{DeBlurGAN}. The GoPro dataset consists of 2,013 and 1,111 non-overlapped training (GoPro-train) and test (GoPro-test) pairs of blurred and sharp images. We show quantitative results in the Table \ref{tab:go_pro_results}. DeblurGAN yields outstanding SSIM number, whereas the proposed DuRN-U is the best in terms of PSNR. Examples of deblurred images are shown in Fig.~\ref{fig:go_pro_vis}. It is observed that the details such as cracks on a stone-fence or numbers written on the car plate are restored well enough to be recognized. 

\myparagraph{Results: Object Detection from Deblurred Images}
In \cite{DeBlurGAN}, the authors evaluated their deblurring method (DeBlurGAN) by applying an object detector to the deblurred images obtained by their method. Following the same procedure and data (Car Dataset), we evaluate our DuRN-U that is trained on the GoPro-train dataset. The Car Dataset contains 1,151 pairs of blurred and sharp images of cars. We employ YOLO v3 \cite{yolov3} trained using the Pascal VOC \cite{voc} for the object detector.
The detection results obtained for the sharp image by the same YOLO v3 detector are utilized as the ground truths used for evaluation. Table \ref{tab:the_car_results} shows quantitative results (measured by mAP), from which it is seen that the proposed DuRN-U outperforms the state-of-the-art DeBlurGAN. Figure \ref{fig:the_car_det} shows examples of detection results on the GoPro-test dataset and Car Dataset. It is observed that DuRN-U can recover details to a certain extent that improves accuracy of detection.

\subsection{Haze Removal}
\label{sec:haze}
\myparagraph{Network Design}
In contrast with previous studies where a CNN is used to explicitly estimate a transmission map that models the effects of haze, we pursue a different strategy, which is to implicitly estimate a transmission map  using an attention mechanism. Our model estimates the dehazed image from an input image in an end-to-end fashion. We design DuRB's for this task by employing 
up-sampling ($up$) implemented using PixelShuffle \cite{pixelshuffle} with a convolutional layer ($ct^{l}_{1}$) in $T^{l}_{1}$ and channel-wise attention ($se$) implemented using SE-ResNet module \cite{senet} with a conv. layer ($ct^{l}_{2}$) in $T^{l}_{2}$. More details are given in the supplementary material. 
\begin{figure}[t]
\centering
\includegraphics[clip, width=\columnwidth]{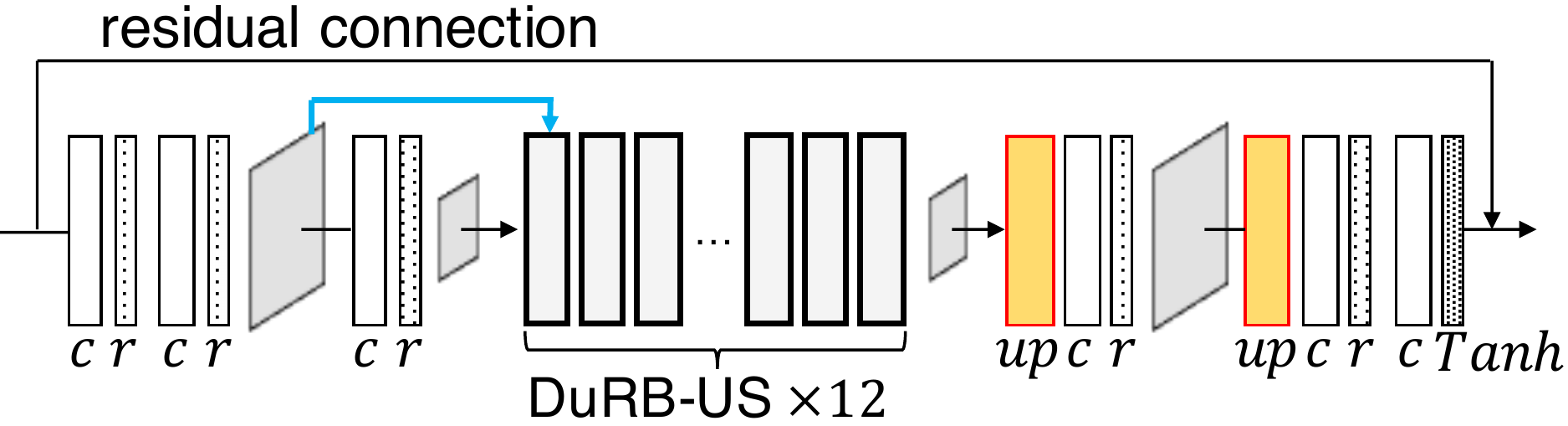}
\caption{DuRN-US: dual residual network with DuRB-US's (up- and down-sampling and channel-wise attention (SE-ResNet Module)) for haze removal. }
\label{fig:DuRN_dehaze}
\vspace{-0.3cm}
\end{figure}
\begin{table}[t]
\centering
\caption{Results for haze removal on Dehaze-TestA dataset and RESIDE-SOTS dataset.}
\vspace{-0.3cm}
\label{tab:haze_zhanghe_result}
\resizebox{\columnwidth}{!}{
\begin{tabular}{cc}
\begin{tabular}{|c|c|}
 \hline
 \multicolumn{2}{|c|}{Dehaze-TestA} \\
 \hline
 He \etal\cite{kaiming_dehaze} &0.8642 \\
 Zhu \etal\cite{zhu_dehaze} &0.8567 \\ 
 Berman \etal\cite{berman_dehaze} &0.7959 \\
 Li \etal\cite{aodnet} &0.8842 \\
 Zhang \etal\cite{dehaze_zhanghe} &{0.9560} \\
 \hline\hline
 \textbf{DuRN-US (ours)} &{\bf 0.9827} \\
 \hline
\end{tabular} &
\begin{tabular}{|c|c|}
 \hline
 \multicolumn{2}{|c|}{RESIDE-SOTS} \\
 \hline 
  Berman \etal\cite{berman_dehaze}& 17.27 / 0.75 \\
  Ren \etal\cite{Ren_dehaze}& 17.57 / 0.81 \\
  Cai \etal\cite{dehazenet}& 21.14 / 0.85 \\
  Li \etal\cite{aodnet}& 19.06 / 0.85 \\
  Ren \etal\cite{GFN}& {22.30} / {0.88} \\
  \hline\hline
  \textbf{DuRN-US (ours)}& {\bf 32.12} / {\bf 0.98} \\
  \hline
\end{tabular} 
\end{tabular}
}
\end{table}
The entire network (named DuRN-US) has an encoder-decoder structure similar to the DuRN-U designed for motion blur removal, as shown in Fig.~\ref{fig:DuRN_dehaze}. We stack 12 DuRB-US's in the middle of the network; the number of channels is 64 for all the layers. In the supplementary material, we demonstrate how our network estimates a transmission map inside its attention mechanisms. For this task, we employed a weighted sum of SSIM and $l_1$ loss for training the DuRN-US.

\begin{figure*}[t]
\centering
\includegraphics[clip, width=\textwidth]{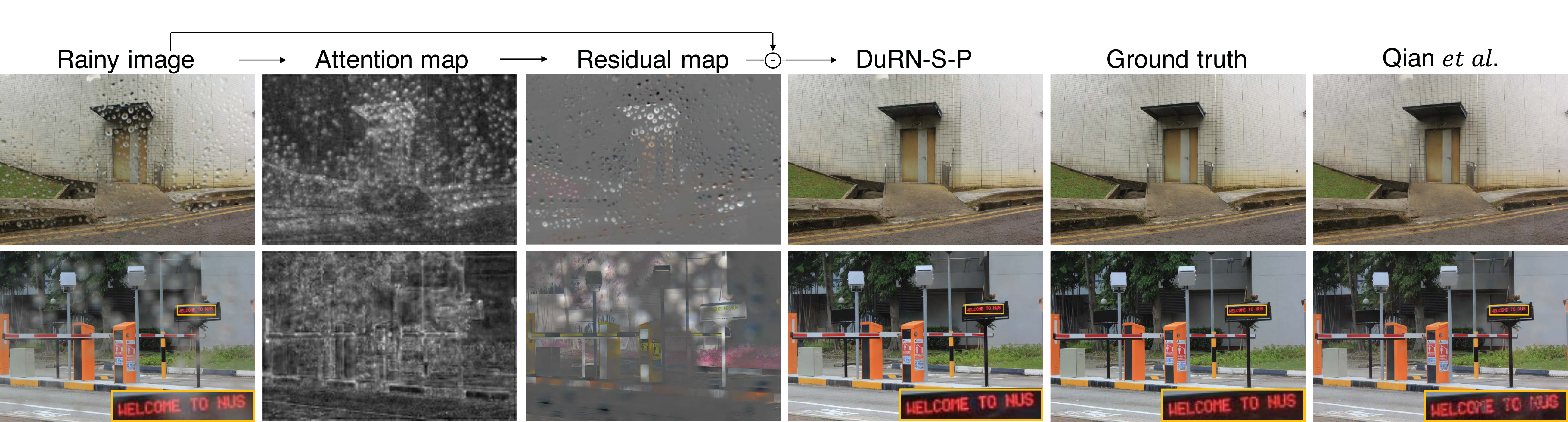}
\caption{Examples of raindrop removal along with internal activation maps of DuRN-S-P. The ``Attention map'' and ``Residual map'' are the outputs of the Attentive-Net and the last $Tanh$ layer shown in Fig.~\ref{fig:DuRB-S-P}; they are normalized for better visibility.}
\label{fig:raindrop_vis}
\vspace{-0.3cm}
\end{figure*}

\myparagraph{Results}
In order to evaluate the proposed DuRN-US, we trained and tested it on two datasets, the Dehaze Dataset and the RESIDE dataset. 
The training and test (Dehaze-TestA) subsets in the Dehaze Dataset consist of 4,000 and 400 non-overlapped samples of indoor scenes, respectively. 
RESIDE contains a training subset of 13,990 samples of indoor scenes and a few test subsets. Following \cite{GFN}, we used a subset SOTS (Synthetic Objective Testing Set) that contains 500 indoor scene samples for evaluation. It should be noted that the state-of-the-art method on the Dehaze Dataset, DCPDN \cite{dehaze_zhanghe}, is trained using i) hazy images, ii) ground truth images, iii) ground truth global atmosphere light , iv) ground truth transmission maps; additionally, its weights are initialized by those of DenseNet \cite{densenet}  pre-trained on the ImageNet\cite{imagenet}. The proposed DuRN-US is trained only using i) and ii).
Table \ref{tab:haze_zhanghe_result} show results on Dehaze-TestA and RESIDE-SOTS datasets, respectively.

Figure \ref{fig:haze_testA} shows examples of the results obtained by the proposed network and others for the same input images. In  sub-figure (A), we show results for two synthesized images produced by the DCPDN (the second best approach in terms of SSIM and PSNR) and our DuRN-US. It is observed that DuRN-US yields better results for these two images. In  sub-figure (B), we show results for two real-world hazy images\footnote{The images are available from https://github.com/rwenqi/GFN-dehazing} produced by two state-of-the-art methods, GFN\cite{GFN} and DCPDN\cite{dehaze_zhanghe}, and by ours. It can be observed that our network yields the most realistic dehazed images.
It is noteworthy that our DuRN-US can properly deal with strong ambient light (sunshine coming behind the girl). See the example in the left-bottom of Fig.~\ref{fig:haze_testA}.

\subsection{Raindrop removal}
\label{sec:raindrop}

\myparagraph{Network Design}
The task can naturally be divided into two stages, that of identifying the regions of raindrops and that of recovering the pixels of the identified regions. The second stage is similar to image inpainting and may not be difficult, as there are a lot of successful methods for image inpainting. Then, the major issue is with the first stage. Following this two-stage approach, the state-of-the-art method \cite{raindrop18}  uses an attentive-recurrent network to produce an attention map that conveys information about raindrops; then, the attention map along with the input image are fed to a convolutional encoder-decoder network to estimate the ground truth image. It also employs adversarial training with a discriminator to make the generated images realistic. 

We show our DuRBs are powerful enough to perform these two-stage computations in a standard feedforward network, if we use properly designed DuRBs in proper positions in the entire network. To be specific, we choose the encoder-decoder structure for the entire network, and in its bottleneck part, we set three DuRB-S's followed by six DuRB-P's. For $ct^{l}_{1}$ in the three DuRB-S's, we use convolution with a $3\times 3$ kernel with decreasing dilation rates, 12, 8, and 6, in the forward direction, aiming to localize raindrops in a coarse-to-fine manner in the three DuRB-S's in a row. For the six DuRB-P's,  we employ the same strategy as in noise removal etc., which is to apply a series of convolution with an increasing receptive field size in the forward direction. We call the entire network DuRN-S-P. For this task, we employed a weighted sum of SSIM and $l_1$ loss for training the DuRN-S-P.

\begin{figure}[t]
\centering
\includegraphics[clip, width=\columnwidth]{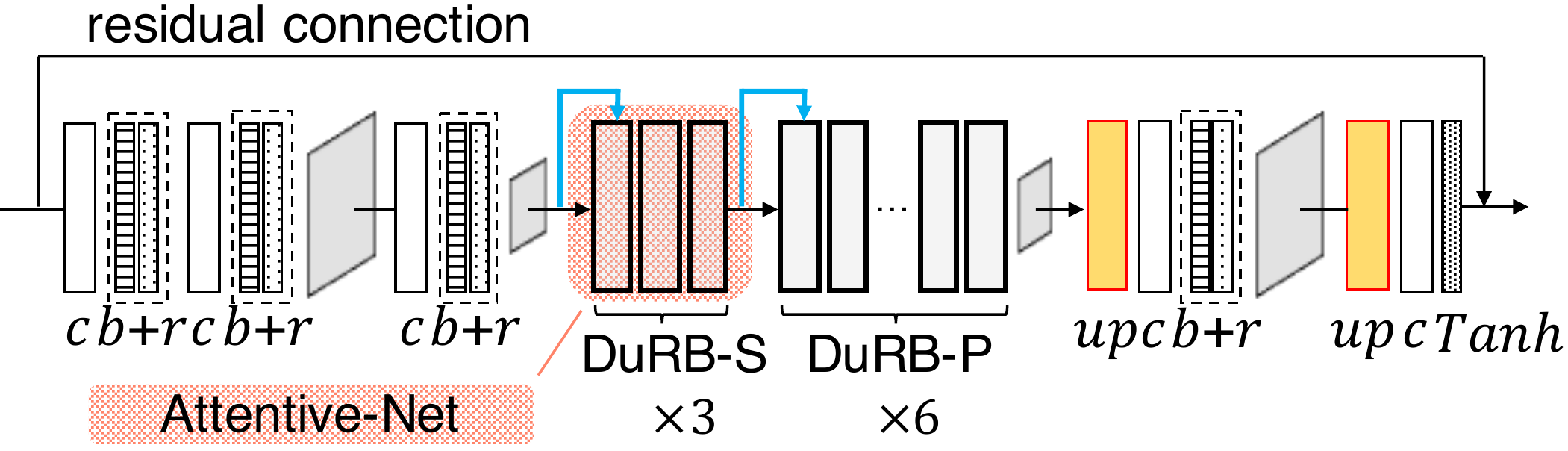}
\caption{DuRN-S-P: Hybrid dual residual network with DuRB-S's and DuRB-P's for raindrop removal.  
}
\label{fig:DuRB-S-P}
\end{figure}

\begin{table}[t]
\centering
\caption{Quantitative result comparison on RainDrop Dataset \cite{raindrop18}.}
\vspace{-0.3cm}
\label{tab:raindrop_result}
\resizebox{.9\linewidth}{!}{
\begin{tabular}{|c|c|c|}
 \hline
   & Qian \etal\cite{raindrop18}  &\textbf{DuRN-S-P (ours)}\\
 \hline
  TestSetA& {\bf 31.51} / {0.9213} & {31.24} / {\bf 0.9259} \\
  TestSetB&  {24.92 / 0.8090}& {\bf 25.32} / {\bf 0.8173} \\
 \hline
\end{tabular}}
\end{table}

\begin{figure*}[t]
\centering
\includegraphics[width=\textwidth]{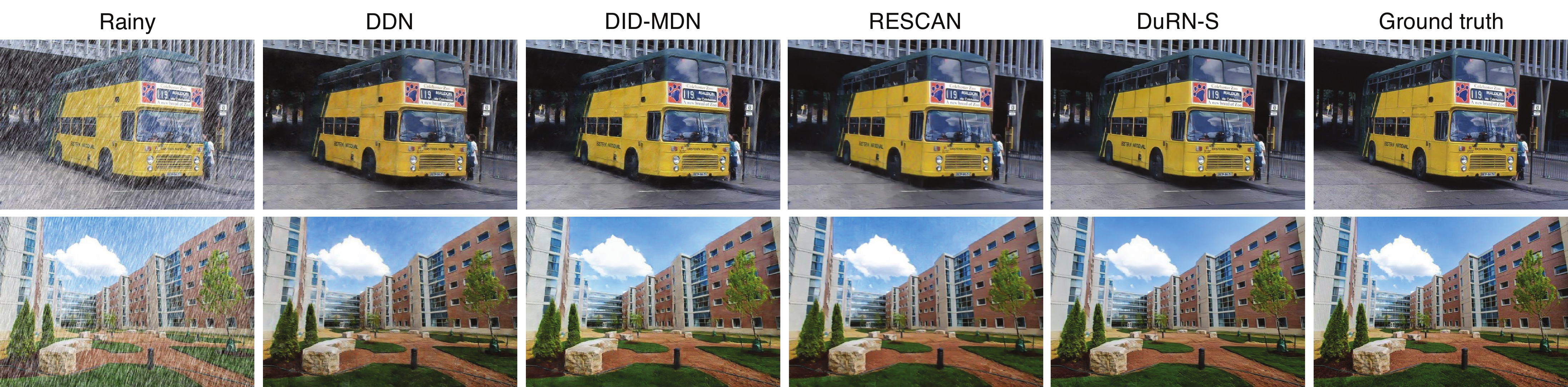}
\caption{Examples of rain-streak removal obtained by four  methods including ours (DuRN-S).}
\label{fig:derain_results}
\vspace{-0.3cm}
\end{figure*}

\myparagraph{Results}
We trained and evaluated the DuRN-S-P on the RainDrop Dataset. It contains 861 training samples and 58/249 test samples called TestSetA/TestSetB. TestSetA is a subset of TestSetB, and is considered to have better alignment\footnote{https://github.com/rui1996/DeRaindrop} than  TestSetB. Table \ref{tab:raindrop_result} shows the results. It is seen that our method outperforms the state-of-the-art method for three out of four combinations of two test sets and two evaluation metrics. It is noteworthy that our method does not use a recurrent network or adversarial training. Figure \ref{fig:raindrop_vis} shows some examples of the results obtained by our method and the method of \cite{raindrop18}. It is seen that the results of our method are visually comparable to the method of \cite{raindrop18}. The ``Attention map'' and ``Residual map'' of Fig.~\ref{fig:raindrop_vis} are the over-channel summation of the output of Attentive-Net and the output of the last $Tanh$ layer, respectively; see Fig.~\ref{fig:DuRB-S-P}.

\subsection{Rain-streak Removal}
\label{sec:rain}

\myparagraph{Network Design}
It is shown in \cite{derain_sota_acm} that the mechanism that selectively weighs 
feature maps using global information works effectively for this task. Borrowing this idea, we employ a channel-wise attention mechanism to perform similar feature weighting. 
The overall design of the network for this task is similar to the DuRN-P designed for Gaussian noise removal. A difference is that we use DuRB-S instead of DuRB-P to use the  attention mechanism. The details are given in the supplementary material. For this task, we employed a weighted sum of SSIM and $l_1$ loss for training the network. 

\myparagraph{Results} 
We tested the proposed network (DuRN-S) on two benchmark datasets, the DDN-Data, which consists of 9,100 training pairs and 4,900 test pairs of rainy and clear images, and the DID-MDN Data, which consists of 12,000 training pairs and 1,200 test pairs. Table \ref{tab:derain_results2} shows the results. Those for the previous methods except RESCAN \cite{rescan} are imported from \cite{derain_sota_acm}. It is seen that the proposed network achieves the best performance. 
Examples of the output images are provided in Fig.~\ref{fig:derain_results}. 

\begin{table}[!t]
\centering
\caption{Results on two de-raining datasets. }
\vspace{-0.3cm}
\label{tab:derain_results2}
\resizebox{.9\linewidth}{!}{
\begin{tabular}{|c|c|c|}
 \hline
   &DDN Data & DID-MDN Data \\
 \hline
   {DDN\cite{xiamen}}& {28.24 / 0.8654} & {23.53} / {0.7057} \\
  {JORDER \cite{jorder}}& {28.72 / 0.8740}& {30.35 / 0.8763} \\
  DID-MDN \cite{derain_zhanghe}& {26.17 / 0.8409}& {28.30} / {0.8707} \\
  RESCAN \cite{rescan} &-/- &32.48 / 0.9096 \\
  NLEDN \cite{derain_sota_acm}& 29.79 / 0.8976 &{33.16} / 0.9192 \\  
  \hline\hline
  \textbf{DuRN-S (ours)}& {\bf 31.30} / {\bf 0.9194} & {\bf 33.21} / {\bf 0.9251} \\
 \hline
\end{tabular}}
\vspace{-0.3cm}
\end{table}

\section{Summary and Discussions}
We have proposed a style of residual connection, dubbed ``dual residual connection'', aiming to exploit the potential of paired operations for image restoration tasks. We have shown the design of a modular block (DuRB) that implements this connection style, which has two containers for the paired operations such that the user can insert any arbitrary operations to them. We have also shown choices of the two operations in the block as well as the entire networks (DuRN) containing a stack of the blocks for five different image restoration tasks. The experimental results obtained using nine datasets show that the proposed approach consistently works better than previous methods. 

\section*{Acknowledgement}
This work was partly supported by JSPS KAKENHI Grant Number JP15H05919,  JST CREST Grant Number JPMJCR14D1, Council for Science, Technology and Innovation (CSTI), Cross-ministerial Strategic Innovation Promotion Program (Infrastructure Maintenance, Renovation and Management ), and the ImPACT Program “Tough Robotics Challenge” of the Council for Science, Technology, and Innovation (Cabinet Office, Government of Japan).

{\small
\bibliographystyle{ieee}
\bibliography{egbib}
}

\begin{table*}
\begin{center}
  {\Large \bf Supplementary material for ``Dual Residual Networks Leveraging \\ the Potential of Paired Operations for Image Restoration'' \par} \vspace*{24pt}
\end{center}
\end{table*}

\setcounter{equation}{0}
\setcounter{figure}{0}
\setcounter{table}{0}
\pagebreak
\appendix

This document provides additional explanations about the experimental setting for each of the five image restoration tasks.

\section{Implementation Details and Additional Results for the Five Tasks}

\subsection{Details of Training Method}
We use the Adam optimizer with $(\beta_1,\beta_2)=(0.9, 0.999)$ and $\epsilon=1.0\times 10^{-8}$ for training all the proposed DuRNs. For loss functions, we use a weighted sum of SSIM and $l_{1}$ loss, specifically, $1.1\times\mbox{SSIM}+0.75\times l_1$, for all the tasks. 
There are two exceptions. One is Gaussian noise removal on the BSD500-grayscale dataset \cite{BSD}, where we use $l_2$ loss. The other is raindrop removal, where we use the same weighted loss for the first 4,000 epochs, and then switch it to 
a single $l_1$ loss for additional 100 epochs. The initial learning rate is set to 0.0001 for all the tasks.    
All the experiments are conducted using PyTorch \cite{pytorch}. Our code and trained models will be made publicly available at {\em https://github.com/liu-vis/DualResidualNetworks}

\subsection{Additional Results}
Additional results for visual quality comparison for the five tasks will be provided with the code in our repository in Github, due to the file size limitation.

\subsection{Noise Removal}
\begin{table}[!b]
\centering
\caption{The specification of $ct^{l}_{1}$ and $ct^{l}_{2}$ for DuRB-P's for noise removal. The ``recep.'' denotes the receptive field of convolution, i.e., 
$\mbox{delation rate}\times\mbox{(kernel size - 1)} + 1$.
}
\vspace{-0.25cm}
\label{tab:noisy_DuRB_p_setting}
\resizebox{\linewidth}{!}{
\begin{tabular}{|c|c|c|c||c|c|c|c|}
 \hline
  layer & kernel & dilation &recep. &layer & kernel & dilation & recep.\\
   \hline
  $ct^{l=1}_{1}$& 5& 1 & 5$\times$5 &$ct^{l=1}_{2}$& 3& 1& 3$\times$3\\
  \hline
  $ct^{l=2}_{1}$& 7& 1 & 7$\times$7 &$ct^{l=2}_{2}$& 5& 1& 5$\times$5\\  
  \hline
  $ct^{l=3}_{1}$& 7& 2 & 13$\times$13 &$ct^{l=3}_{2}$& 5& 1& 5$\times$5\\
 \hline
  $ct^{l=4}_{1}$& 11& 2& 21$\times$21 &$ct^{l=4}_{2}$& 7& 1& 7$\times$7\\
 \hline
  $ct^{l=5}_{1}$& 11& 1& 11$\times$11 &$ct^{l=5}_{2}$& 5& 1& 5$\times$5\\
 \hline
  $ct^{l=6}_{1}$& 11& 3& 31$\times$31 &$ct^{l=6}_{2}$& 7& 1& 7$\times$7\\
 \hline
\end{tabular}}
\end{table}

\myparagraph{Specification of $ct^{l}_{1}$ and $ct^{l}_{2}$} 
We show the specification of $ct^{l}_{1}$ and $ct^{l}_{2}$ for each DuRB-P in Table \ref{tab:noisy_DuRB_p_setting}, in which $l (=1,\ldots,6)$ denotes the block-id of a DuRB; the ``recep.'' denotes the receptive field of convolution. It is observed that the paired convolution has a large- and small- receptive field for each DuRB-P (see each row in the table), and the size of the receptive fields of $ct^{l}_{1}$ and $ct^{l}_{2}$ increases with $l$ with an exception at $l=5$, which is to avoid too large a receptive field. By this design we intend to make each block look at the input image at an increasing scale with layers in the forward direction.

\myparagraph{Experimental Setting for Gaussian Noise Removal} In training, we set batch size $=100$. Each input image in a batch is obtained by randomly cropping a $64\times64$ region from an original training noisy image.
We exactly followed the procedure of \cite{god_suganuma} to generate noisy images for training our network.

\myparagraph{Experimental Setting for Real-World Noise Removal} In training, we randomly select 30 out of 40 pairs of a high resolution noisy image and a mean image (used as ground truth) for constructing the training dataset.
We set input patch size $=128\times128$, and use 30 patches (each of which is randomly cropped from a different training image) to create one batch. 
To test the CNNs including ours and the baselines, we use the remaining 10 image pairs;  specifically, we randomly crop ten $512\times512$ patches from each of them, yielding 100 patches that are used for the test.

\subsection{Motion Blur Removal}
\begin{table}[!h]
\centering
\caption{The specification of $ct^{l}_{1}$ for DuRB-U's for motion blur removal.}
\vspace{-0.25cm}
\label{tab:blur_ct_setting}
\resizebox{\linewidth}{!}{
\begin{tabular}{|c|c|c|c||c|c|c|c|}
 \hline
  layer & kernel & dilation &recep. &layer & kernel& dilation &recep. \\
   \hline
  $ct^{l=1}_{1}$& 3& 3& 7&$ct^{l=4}_{1}$& 7& 1& 7\\
  \hline
  $ct^{l=2}_{1}$& 7& 1& 7&$ct^{l=5}_{1}$& 3& 2& 5\\  
  \hline
  $ct^{l=3}_{1}$& 3& 3& 7&$ct^{l=6}_{1}$& 5& 1& 5\\
 \hline
\end{tabular}}
\end{table}

\myparagraph{Specification of $ct^{l}_{1}$ and $ct^{l}_{2}$} 
The specification of $ct^{l}_{1}$ is shown in Table \ref{tab:blur_ct_setting}. For $ct^{l}_{2}$, we use an identical configuration, kernel size $=3\times3$, dilation rate $=1$ and stride $=2$, for all DuRB-U's. We intend to simply perform down-sampling with $ct^l_2$.

\myparagraph{Experimental Setting on GoPro Dataset}
In training, we set batch size $=10$. Each input image in a batch is obtained by randomly cropping a $256\times256$ patch from the re-sized version ($640\times360$) of an original training image of size $1280\times720$.
In testing, we use the re-sized version ($640\times360$) of the original test images of size $1280\times720$ as in training.

\myparagraph{Experimental Setting on Car Dataset}
The Car dataset was used only for evaluation. We down-scale the blur images from their original size $720\times720$ to $360\times360$ and input them to the DuRN-U trained using GoPro-train dataset for de-blurring. The result is then up-scaled to $700\times700$ and fed into YOLOv3.


\subsection{Haze Removal}
\begin{table}[!t]
\centering
\caption{The specification of $ct^{l}_{1}$ for DuRB-US's for haze removal.}
\vspace{-0.25cm}
\label{tab:DuRB-US-ct1}
\resizebox{\linewidth}{!}{
\begin{tabular}{|c|c|c|c||c|c|c|c|}
 \hline
  layer & kernel & dilation& recep.& layer &kernel & dilation &recep. \\
   \hline
  $ct^{l=1}_{1}$& 5& 1& 5&$ct^{l=7}_{1}$& 11& 1& 11\\
  \hline
  $ct^{l=2}_{1}$& 5& 1& 5&$ct^{l=8}_{1}$& 11& 1& 11\\  
  \hline
  $ct^{l=3}_{1}$& 7& 1& 7&$ct^{l=9}_{1}$& 11& 1& 11\\
  \hline
  $ct^{l=4}_{1}$& 7& 1& 7&$ct^{l=10}_{1}$& 11& 1& 11\\
  \hline
  $ct^{l=5}_{1}$& 11& 1& 11&$ct^{l=11}_{1}$& 11& 1& 11\\
  \hline
  $ct^{l=6}_{1}$& 11& 1& 11&$ct^{l=12}_{1}$& 11& 1& 11\\
 \hline
\end{tabular}}
\end{table}
\myparagraph{Specification of $ct^{l}_{1}$ and $ct^{l}_{2}$} The specification of $ct^{l}_{1}$ is shown in Table \ref{tab:DuRB-US-ct1}. For $ct^{l}_{2}$, we use an identical configuration, i.e., kernel size $=3\times3$, dilation rate $=1$ and stride $=2$, for all the DuRB-US's. We intend to simply perform down-sampling with $ct^l_2$.

\myparagraph{Experimental Setting on Dehaze Dataset}
In training, we set batch size $=20$. Each input image in a batch is obtained by randomly cropping a $256\times256$ region from an original training image of size $512\times512$. 

\myparagraph{Experimental Setting on RESIDE}
In training, we set batch size $=48$. Each input image in a batch is obtained by randomly cropping a $256\times256$ region from an original image of size $620\times460$. 

\begin{figure*}[t]
\centering
\includegraphics[clip, width=\textwidth]{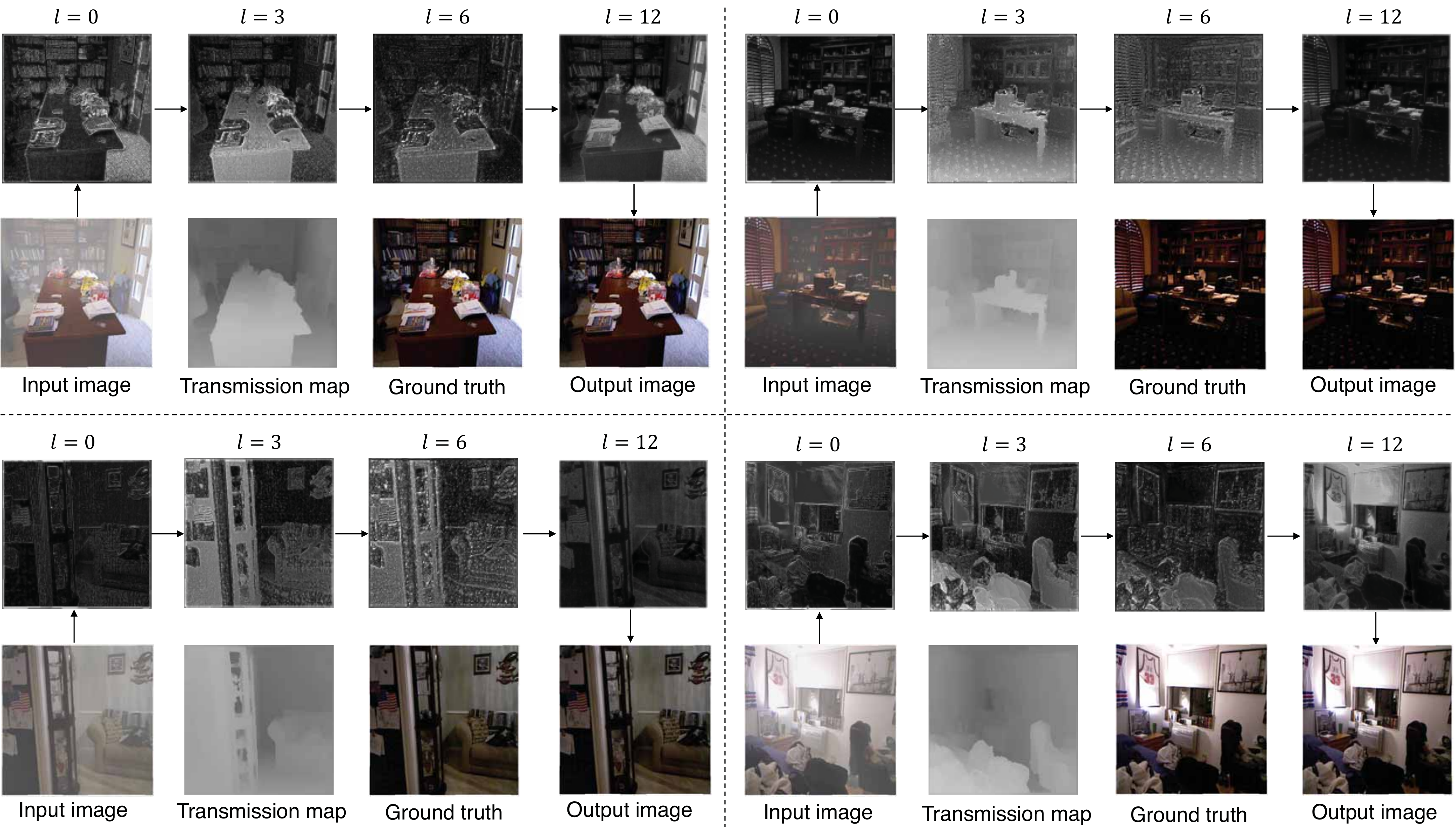}
\caption{Visualization of internal activation maps of the DuRN-US.
}
\label{fig:dehaze_inside}
\vspace{-0.3cm}
\end{figure*}

\myparagraph{Visualization of Internal Layer Activation}
Figure \ref{fig:dehaze_inside} shows activation maps of several chosen blocks (i.e., DuRB-US's) in the network for different input images. They are the sums in the channel dimension of activation maps of the input to the first DuRB-US ($l=0$), and of the output from the third ($l=3$), sixth ($l=6$), and twelfth ($l=12$) DuRB-US's. It is seen that the DuRN-US computes a map that looks similar to transmission map at around $l=3$. 


\subsection{Raindrop Removal}
\myparagraph{Specification of $ct^{l}_{1}$ and $ct^{l}_{2}$} The specification of $ct^{l}_{1}$ for the three DuRB-S's and the six DuRB-P's is shown in Table \ref{tab:DuRN-S-P_ct1}. For $ct^{l}_{2}$, we use an identical configuration, kernel size $=3\times3$ and dilation rate $=1$, for all the DuRB-S's, and use an identical configuration, kernel size $=5\times5$ and  dilation rate $=1$, for all the DuRB-P's. 

\myparagraph{Experimental Setting on RainDrop Dataset}
In training, we set batch size $=24$. Each input image in a batch is obtained by randomly cropping a $256\times256$ region from the original image of size $720\times480$. 
As mentioned before, we train the network $1.1\times\mbox{SSIM}+0.75\times l_1$ using the loss for 4,000 epochs, and then switch the loss to $l_1$ alone, training the network for additional 100 epochs. We did this for faster converging. 


\begin{table}[t]
\centering
\caption{The specification of $ct^{l}_{1}$ for DuRB-S's and DuRB-P's of the DuRN-S-P for raindrop removal.}
\vspace{-0.25cm}
\label{tab:DuRN-S-P_ct1}
\resizebox{\linewidth}{!}{
\begin{tabular}{cc}
\begin{tabular}[t]{|c|c|c|c|}
 \hline
  \multicolumn{4}{|c|}{DuRB-S} \\
 \hline 
  layer & kernel & dilation& recep. \\
   \hline
  $ct^{l=1}_{1}$& 3& 12& 25 \\
  \hline
  $ct^{l=2}_{1}$& 3& 8& 17 \\  
  \hline
  $ct^{l=3}_{1}$& 3& 6& 13 \\
  \hline
\end{tabular}&
\begin{tabular}[t]{|c|c|c|c|}
 \hline
  \multicolumn{4}{|c|}{DuRB-P} \\
 \hline
  layer & kernel & dilation& recep. \\
   \hline
  $ct^{l=1}_{1}$& 3& 2& 5 \\
  \hline
  $ct^{l=2}_{1}$& 5& 1& 5 \\  
  \hline
  $ct^{l=3}_{1}$& 3& 3& 7 \\
  \hline
  $ct^{l=3}_{1}$& 7& 1& 7 \\
  \hline
  $ct^{l=3}_{1}$& 3& 4& 9 \\
  \hline
  $ct^{l=3}_{1}$& 7& 1& 7 \\
  \hline
\end{tabular}
\end{tabular}
}
\end{table}

\subsection{Rain-streak Removal}

\myparagraph{Specification of $ct^{l}_{1}$ and $ct^{l}_{2}$} We use the same configuration as noise removal. See Table.~\ref{tab:noisy_DuRB_p_setting}. Note that we use DuRB-S for this task. 

\myparagraph{Experimental Setting on DDN Data} To train the DuRN-S, we set batch size $=40$. Each input image in a batch is obtained by randomly cropping a $64\times64$ region from an original training image.

\myparagraph{Experimental Setting on DID-MDN Data} 
In training, we set batch size $=80$. Each input image in a batch is obtained by randomly cropping a $64\times64$ region from an original training image. 


\begin{table*}[t]
\centering
 \caption{Performance (PSNR/SSIM) of the four versions  of DuRBs (i.e., -P, -U, -US, and -S) on different task.}
 \label{tab:DuRB_fivetasks}
 \small
 \begin{tabular}[t]{|c|c|c|c|c|c|}
 \hline
  & Real-noise  & Motion blur  & Haze  & Raindrop & Rain-streak\\
 \hline
  DuRB-P& {\bf 36.83} / {\bf 0.9635}& 29.40 / 0.8996& 29.33 / 0.9761 &  24.69 / 0.8067& 32.88 / 0.9214\\
  DuRB-U& 36.63 / 0.9600 & 29.90 / 0.9100& 30.79 / 0.9800 & 24.30 / 0.8067& 33.00 / {\bf 0.9265}\\
  DuRB-US& 36.61 / 0.9591 & {\bf 29.96} / {\bf 0.9101} & {\bf 32.60} / {\bf 0.9827} & 22.72 / 0.7254& 32.84 / 0.9238\\
  DuRB-S& 36.82 / 0.9629 & 29.55 / 0.9023 & 31.81 / 0.9792 & {\bf 25.13} / {\bf 0.8134} & {\bf 33.21} / 0.9251\\
 \hline
 \end{tabular}
\end{table*}

\subsection{Performance of DuRBs on Non-target Tasks}
We have presented the four versions of DuRB, each of which is designed for a single task. To verify the effectiveness of the design choices, we examine the performance of each DuRB on its non-target tasks. Specifically, we evaluate the performance of every combination of the four versions 
of DuRB and the five tasks. For noise, motion blur, haze, raindrop and rain-streak removal, we train and test networks consisting of each version of DuRB on Real-World Noisy Image Dataset, GoPro Dataset, Dehaze Dataset, RainDrop Dataset and DID-MDN Data.
The results are shown in Table \ref{tab:DuRB_fivetasks}.
It is seen that in general, each DuRB yields the best performance for the task to which it was designed. 
For motion blur removal, DuRB-US performs comparably well or even slightly better than DuRB-U, which is our primary design for the task. We think this is reasonable, as DuRB-US contains the same paired operation as DuRB-U (i.e., up- and down-sampling), contributing to the good performance. Their performance gap is almost negligible and thus DuRB-U is a better choice, considering its efficiency. 





\end{document}